\documentclass[lettersize,journal]{IEEEtran}
\usepackage{amsmath,amsfonts}
\usepackage{algorithmic}
\usepackage{algorithm}
\usepackage{array}
\usepackage[caption=false,font=normalsize,labelfont=sf,textfont=sf]{subfig}
\usepackage{textcomp}
\usepackage{stfloats}
\usepackage{url}
\usepackage{verbatim}
\usepackage{graphicx}
\usepackage{cite}
\usepackage{amssymb}
\usepackage{booktabs}
\usepackage{multirow}
\usepackage{bbding}
\usepackage{caption}
\begin{document}

\title{ESGaussianFace: Emotional and Stylized Audio-Driven Facial Animation via 3D Gaussian Splatting}

\author{Chuhang~Ma, Shuai~Tan, Ye~Pan$^{\ast}$, Jiaolong~Yang and Xin~Tong

\thanks{This paper was produced by the IEEE Publication Technology Group. They are in Piscataway, NJ.}
\thanks{Manuscript received April 19, 2021; revised August 16, 2021.}}

\markboth{}%
{Shell \MakeLowercase{\textit{et al.}}: A Sample Article Using IEEEtran.cls for IEEE Journals}

\IEEEpubid{}

\twocolumn[{%
\renewcommand\twocolumn[1][]{#1}%
\maketitle
\begin{center}
    \centering
    \captionsetup{type=figure}
    \includegraphics[width=\textwidth]{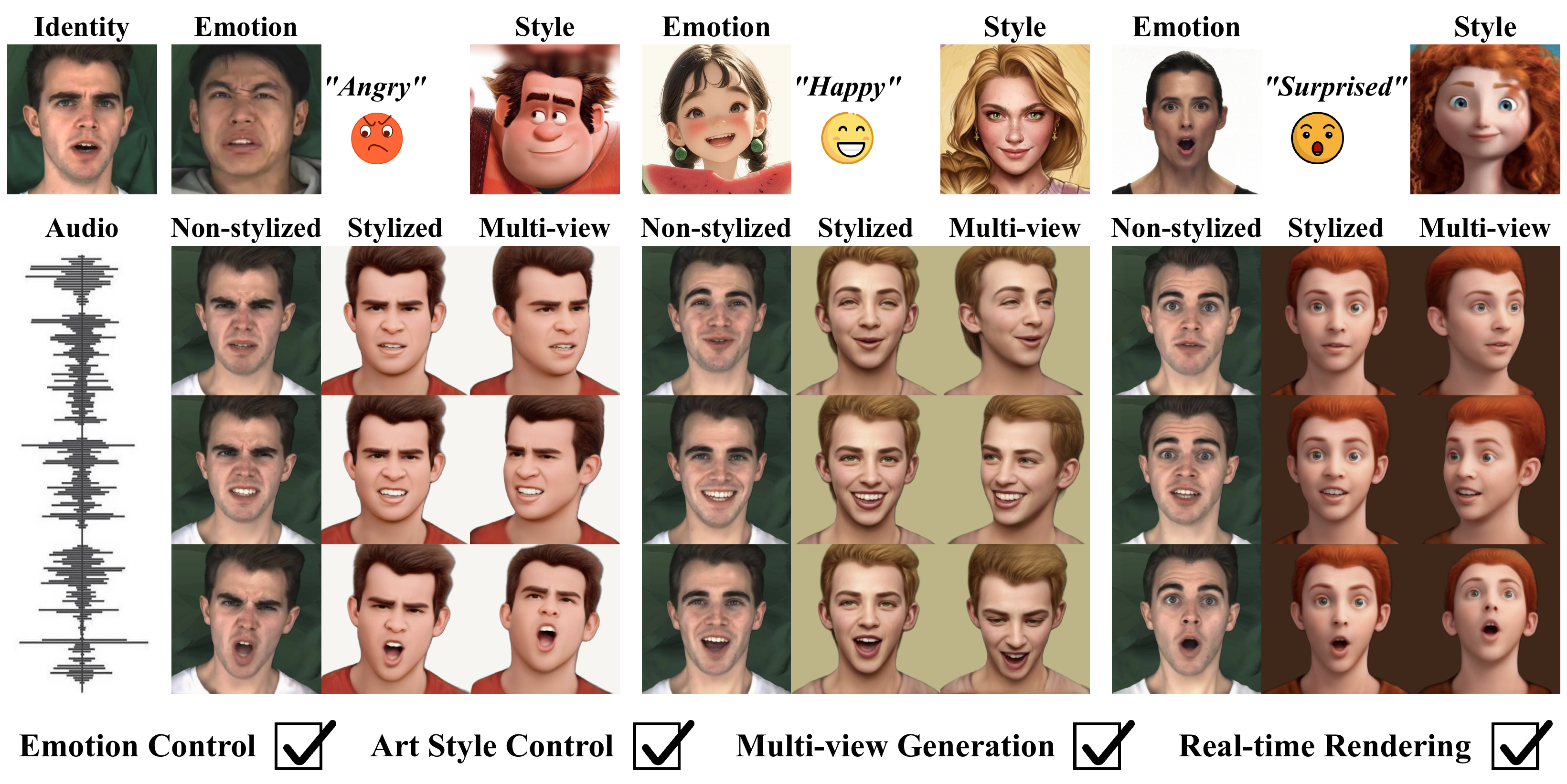}
    \captionof{figure}{We propose an Audio-Driven Facial Animation (ADFA) framework, named \textit{ESGaussianFace}. This method is based on 3D Gaussian Splatting and supports the generation of talking heads with \textit{diverse emotions art styles}. In contrast to prior ADFA approaches, our framework enables the training of a single model capable of generating \textit{high-precision}, \textit{multi-view consistent} videos with varying emotions \textit{in real time}.}
    \label{introd}
\end{center}%
}]

\renewcommand{\thefootnote}{}
\footnotetext{\rule{5cm}{0.4pt}}
\footnotetext{C. Ma, S. Tan and Y. Pan are with JHC \& AI Institute, Shanghai Jiao Tong University, Shanghai, China, E-mail:\{mch3148300494, tanshuai0219, whitneypanye\}@sjtu.edu.cn. J. Yang and X. Tong are with Microsoft Research Asia. Corresponding author: Ye Pan.}

\begin{abstract}
Most current audio-driven facial animation research primarily focuses on generating videos with neutral emotions. While some studies have addressed the generation of facial videos driven by emotional audio, efficiently generating high-quality talking head videos that integrate both emotional expressions and style features remains a significant challenge. In this paper, we propose ESGaussianFace, an innovative framework for emotional and stylized audio-driven facial animation. Our approach leverages 3D Gaussian Splatting to reconstruct 3D scenes and render videos, ensuring efficient generation of 3D consistent results. We propose an emotion-audio-guided spatial attention method that effectively integrates emotion features with audio content features. Through emotion-guided attention, the model is able to reconstruct facial details across different emotional states more accurately. To achieve emotional and stylized deformations of the 3D Gaussian points through emotion and style features, we introduce two 3D Gaussian deformation predictors. Futhermore, we propose a multi-stage training strategy, enabling the step-by-step learning of the character's lip movements, emotional variations, and style features. Our generated results exhibit high efficiency, high quality, and 3D consistency. Extensive experimental results demonstrate that our method outperforms existing state-of-the-art techniques in terms of lip movement accuracy, expression variation, and style feature expressiveness.
\end{abstract}


\section{Introduction}
\IEEEPARstart{A}{udio-Driven} Facial Animation (ADFA) generates facial animations for specific characters based on a given audio segment. This task has widespread applications in digital humans, virtual reality, and various other industrial domains. Most existing ADFA methods primarily focus on neutral emotions \cite{zhou2020makelttalk, prajwal2020lip, wang2021audio2head}, with only a few \cite{ji2022eamm, tan2023emmn, ma2023dreamtalk, gan2023efficient} using audio with different emotions to generate talking head videos. However, due to the lack of constraints from a 3D scene, the videos produced by these methods lack 3D consistency and cannot achieve multi-view rendering. In recent years, Neural Radiance Fields (NeRF) \cite{mildenhall2021nerf} have achieved significant breakthroughs in 3D reconstruction by modeling 3D scenes using implicit functions. However, NeRF suffers from slow rendering speeds, limiting its ability to achieve real-time rendering. 

3D Gaussian Splatting (3DGS) \cite{kerbl20233d} offers a promising solution to address this limitation. 3DGS is a 3D reconstruction method that enables high-speed rendering through explicit point-based 3D scene
representation and a highly parallel workflow, allowing for near-real-time rendering while maintaining visual quality. Consequently, many ADFA approaches \cite{dhamo2023headgas, chen2023monogaussianavatar} have begun to use 3DGS to model heads of specific characters. However, these efforts primarily focus on ADFA with neutral emotion. Although EmoTalkingGaussian \cite{cha2025emotalkinggaussian} is capable of generating talking heads with predefined emotion labels, generating ADFA videos that integrate \textit{3D consistency}, \textit{emotional expression} and \textit{style features} from an emotional image and a stylized avatar has not yet been explored.

\begin{figure}[t]
\centering
\includegraphics[width=0.94\columnwidth]{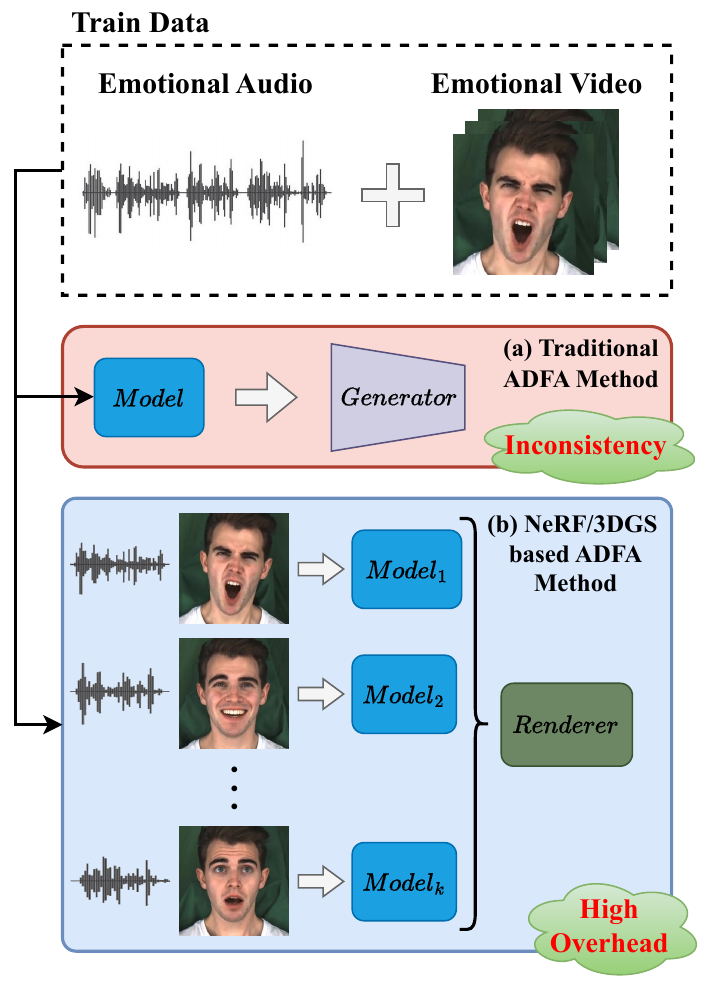} 
\caption{(a) Traditional ADFA methods and (b) NeRF/3DGS-based ADFA approaches struggle to handle multi-emotion talking head generation tasks.}
\label{introd_half}
\end{figure}

We propose a novel framework named \textit{ESGaussianFace}, designed to generate both emotional and stylized talking head videos efficiently. Fig. \ref{introd_half} illustrates the limitations of current state-of-the-art ADFA methods in handling multi-emotion ADFA tasks. Traditional ADFA methods \cite{zhou2020makelttalk, prajwal2020lip}, depicted in Fig. \ref{introd_half} (a), often suffer from a lack of 3D consistency, resulting in inaccurate reconstructions of facial orientation and pose. NeRF/3DGS-based ADFA methods \cite{guo2021ad, cho2024gaussiantalker}, shown in Fig. \ref{introd_half} (b), are proficient at capturing 3D features but struggle with audios that contain mixed emotions. They typically require training multiple models for different emotions, leading to significant time and memory overhead. In contrast, our ESGaussianFace efficiently generates videos for a wide range of emotions using a single model. Moreover, ESGaussianFace allows for the seamless integration of any artistic avatar's style into emotional videos. During training, ESGaussianFace takes driving audios, emotional videos, and style images as inputs. These inputs provide the character's lip movements, emotional expressions, and style features, respectively. During inference, any emotional video or image can be used as the emotion source to extract emotion features. 

To realize the aforementioned advantages, our ESGaussianFace is structured into three modules: triplane-based 3D Gaussian generator, audio-visual feature extraction and fusion module, and ESGaussian deformation prediction module. The \textit{triplane-based 3D Gaussian generator} focuses on generating the Gaussian parameters for the canonical face. We employ a multi-resolution triplane to encode spatial information form a standard 3D head, and a triplane decoder to generate the canonical 3D Gaussian parameters. 

Our goal is to train a 3D Gaussian model capable of accurately generating the target avatar's lip movements and emotional expressions. The former is primarily driven by the input audio, while the latter is controlled by a facial image representing a specific emotion. To achieve this, we extract content features from the audio and employ a 3D Morphable Model \cite{blanz2023morphable} to capture the facial expression coefficients as emotion features in \textit{audio-visual feature extraction and fusion module}. However, efficiently integrating these two feature types to precisely control the 3D Gaussian deformation presents a significant challenge. We hope that these two features can dynamically influence distinct facial regions and control the deformation of 3D Gaussian points within these regions. For this purpose, we propose an \textit{emotion-audio-guided spatial attention module}, which consists of two primary cross-attention layers. The audio-guided attention layer captures the influence of the audio content features on the mouth region, while the emotion-guided attention layer controls the deformation of various facial regions under the guidance of emotions. The module leverages spatial attention to effectively integrate content and emotion features, enabling precise guidance of the 3D Gaussian parameter deformations.

To further generate emotional and stylized talking head videos, we incorporate the \textit{ESGaussian deformation prediction module}. We combine features output by the previous module with the embeddings of 3D point positions to learn more precise emotional variations. Futhermore, we extract style encodings to capture the stylistic attributes of a specific artistic avatar. These features guide the deformation predictor in generating Gaussian deformations. Training a model to directly accomplish such a complex task is challenging. To overcome this, we propose a \textit{multi-stage training strategy}. First, the neutral stage enables the model to predict lip movements under neutral emotion. In the emotion stage, the model learns how different emotions influence the Gaussian parameters, based on the learned lip movements. Finally, to achieve the stylized deformation, we introduce a stylization stage. 

Overall, ESGaussianFace can generate emotional and stylized talking head videos, merging realism, accuracy, and high efficiency. Experimental results show the superiority of our method compared to state-of-the-art methods. The main contributions of this work can be summarized as follows:

\begin{itemize}
    \item We propose a novel framework for tackling the ADFA task that combines both emotion and style features. To the best of our knowledge, we are the first to achieve this task while ensuring efficiency, accuracy, and multi-view generation.
    \item We introduce an emotion-audio-guided spatial attention method to accurately learn the influence of both the audio and the emotion on dynamic spatial points. This method enables the precise prediction of Gaussian deformations.
    \item We design a multi-stage training strategy, consisting of three stages, to train the model. This strategy enables the model to generate more accurate and stable videos. We also introduce several novel loss functions for this task.
\end{itemize}

\section{Related Work}
\subsection{Audio-Driven Facial Animation in 2D Pixel Space}

Audio-Driven Facial Animation (ADFA) tasks involve creating facial animation videos from an input audio clip. Early approaches primarily employ CNN-based encoder-decoder architectures \cite{kumar2017obamanet, jamaludin2019you} or adversarial networks \cite{goodfellow2014generative, chen2019hierarchical, zhou2019talking, prajwal2020lip, yu2020multimodal, das2020speech, wang2021audio2head, sun2021speech2talking, yin2022styleheat} to generate talking head videos. However, they often neglect facial structural features, leading to significant distortions in the generated images. To address this issue, several techniques \cite{zhou2018visemenet, zakharov2019few, lu2021live, zhou2020makelttalk} enhance model accuracy by utilizing facial landmarks as an additional control mechanism. Furthermore, integrating the 3D Morphable Model (3DMM) \cite{blanz2023morphable, taylor2017deep, suwajanakorn2017synthesizing, thies2020neural, yi2020audio, ren2021pirenderer, zhang2023sadtalker} and 3D blendshape face model \cite{karras2017audio, pham2017speech, pham2018end, cudeiro2019capture, tzirakis2020synthesising, richard2021meshtalk, peng2023selftalk} into ADFA systemsproduces more realistic and accurate videos. However, these works primarily focus on faces displaying neutral emotions.

Currently, few studies address the generation of talking head videos with varying facial emotions. MEAD \cite{wang2020mead} and RAVDESS \cite{livingstone2018ryerson} datasets provide valuable emotional audio-visual data with high quality. Several efforts \cite{eskimez2021speech, liang2022expressive, sinha2022emotion, ji2022eamm, ma2023styletalk, tan2023emmn, sun2023continuously, peng2023emotalk, gan2023efficient, ma2023dreamtalk, he2024emotalk3d, xu2024vasa, tan2024say, tan2024edtalk,ma2025goes,tan2025edtalk++,tan2025animate++,ji2025sport, tan2024flowvqtalker, witzig2024emospacetime,Mimir2025,tan2025SynMotion,AnimateX2025,tan2025fixtalk,pan2023emotional, pan2024expressive, pan2025vasa} have developed methods for generating emotional talking head videos. These methods derive emotion features from diverse sources. For instance, EAT \cite{gan2023efficient} utilizes discrete emotion labels, while DreamTalk \cite{ma2023dreamtalk} relies on emotional images. However, they often fall short in managing facial aspects such as orientation and pose, and they cannot render multi-view videos.

\subsection{Audio-Driven Facial Animation via Neural Rendering}

Recently, Neural Radiance Fields (NeRF) \cite{mildenhall2021nerf} have shown exceptional performance in rendering complex 3D scenes. As a result, many ADFA methods \cite{guo2021ad, liu2022semantic, shen2022learning, yao2022dfa, tang2022real, muller2022instant, li2023efficient, yu2023nofa} have incorporated NeRF into their frameworks. For instance, AD-NeRF \cite{guo2021ad} innovatively separates facial generation into head and torso stages, leveraging NeRF's implicit representation to achieve high-quality rendering. However, a major limitation of NeRF is its slow rendering speed, presenting a challenge in balancing efficiency and accuracy in ADFA.

The advent of 3D Gaussian Splatting (3DGS) \cite{kerbl20233d} has significantly addressed this issue by offering both improved rendering quality and faster speeds compared to NeRF. Recent studies in 3D facial animation have begun to adopt 3DGS for its high precision and efficiency. HeadGas \cite{dhamo2023headgas} is pioneering in applying 3DGS into 3D facial reconstruction. MonoGaussianAvatar \cite{chen2023monogaussianavatar} utilizes linear blend skinning to map the 3D points from the canonical space to the deformed space, enabling effective ADFA. Currently, numerous studies \cite{qian2024gaussianavatars, zhou2024headstudio, yu2024gaussiantalker, wang2023gaussianhead, li2024talkinggaussian} integrate the FLAME model \cite{li2017learning} and parameterized tri-plane \cite{Chan2022} into 3DGS methods, yielding more accurate outcomes. For instance, GaussianTalker \cite{cho2024gaussiantalker} employs a multi-resolution tri-plane to represent canonical facial shapes and uses a spatial-audio attention module to predict 3D Gaussian deformation. EmoTalkingGaussian \cite{cha2025emotalkinggaussian} utilizes predefined emotion labels and predicts Gaussian parameter deformations to infuse emotions into talking heads. However, no existing method has yet explored facial stylization based on 3DGS.

\subsection{Style Transfer on Facial Images and Videos}
Facial style transfer is a well-explored research area. DualStyleGAN \cite{yang2022pastiche} employs the pSp encoder \cite{richardson2021encoding} to extract style features from any style image and integrates them with structural features from facial images. Utilizing StyleGAN \cite{karras2019style}, this approach generates high-quality stylized images based on the combined features. VToonify \cite{yang2022Vtoonify} extends this capability to video, enabling the creation of high-precision stylized videos. In ADFA domain, Style$^2$Talker \cite{tan2024style2talker} extracts style features from style images and controls emotions using a diffusion model. However, it is constrained by its 2D generation framework. Although a few studies \cite{men20243dtoonify, shao2024control4d, chen2025revealing} have explored 3D facial style transfer, no existing work is capable of achieving 3D facial animation while simultaneously enabling control over arbitrary emotion and style. In contrast, our method utilizes 3DGS to achieve ADFA with various emotional expressions and styles efficiently.

\begin{figure*}[t]
\centering
\includegraphics[width=\textwidth]{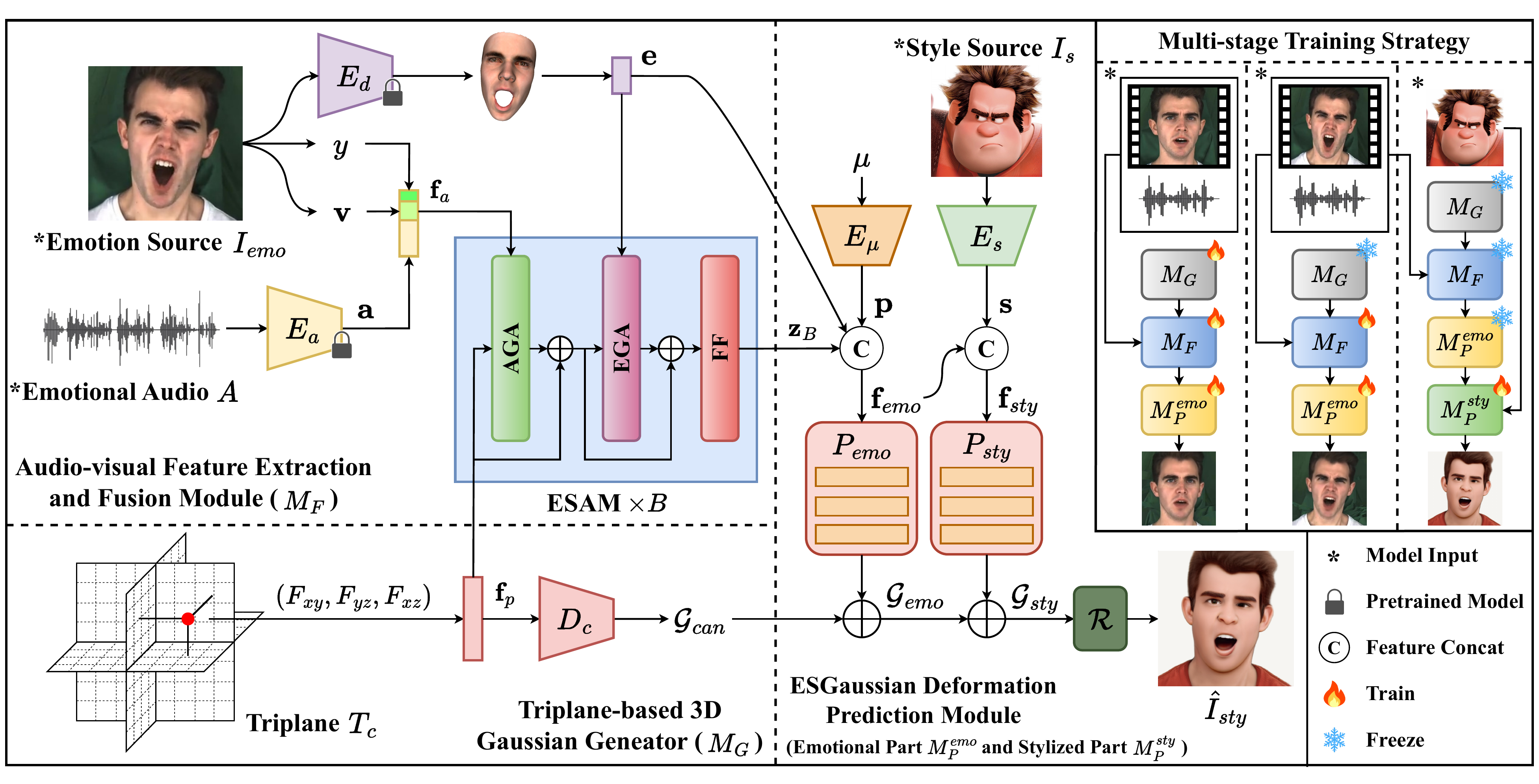} 
\caption{The overview of our proposed \textit{ESGaussianFace} model. During inference, we initialize the Gaussian parameters of the canonical face with the \textit{Triplane-based 3D Gaussian Generator} (Sec. \ref{mod1}). The \textit{Audio-Visual Feature Extraction and Fusion Module} (Sec. \ref{mod2}) extracts content and emotion features from the audio and emotional source separately, then combines them through the Emotion-audio-guided Spatial Attention Module (ESAM). The resulting fused features are subsequently input into the \textit{ESGaussian Deformation Prediction Module} (Sec. \ref{mod3}), which predicts the emotional and stylized deformations of the Gaussian parameters. For training, a \textit{Multi-stage Training Strategy} (Sec. \ref{mod4}) is adopted, with the model learning lip movements, emotional expressions, and style features in three stages.} 
\label{method}
\end{figure*}

\section{Preliminary: 3D Gaussian Splatting}

3D Gaussian Splatting (3DGS) \cite{kerbl20233d} can learn an explicit 3D representation from given images and camera parameters, enabling the reconstruction of stable scenes by a series of 3D Gaussian splats. Typically, a Gaussian splat is defined by its center position $\mathbf{\mu} \in \mathbb{R}^3$, scaling factor $\mathbf{s} \in \mathbb{R}^3$, rotation quaternion $\mathbf{q} \in \mathbb{R}^4$, $k$-degree spherical harmonics coefficients $\mathbf{sh} \in \mathbb{R}^{3(k+1)^2}$, and opacity value $\alpha \in \mathbb{R}$. Therefore, a Gaussian splat can be described as $\mathcal{G} = \{\mathbf{\mu}, \mathbf{s}, \mathbf{q}, \mathbf{sh}, \alpha\}$. Each 3D Gaussian can be computed as follows:
\begin{equation}
    g(\mathbf{x}) = e^{-\frac{1}{2}(\mathbf{x}-\mathbf{\mu})^T\mathbf{\Sigma}^{-1}(\mathbf{x}-\mathbf{\mu})}.
\end{equation}
The semi-definite covariance matrix $\mathbf{\Sigma}$ can be computed from a scaling matrix $\mathbf{S}$ and a rotation matrix $\mathbf{R}$, defined by $\mathbf{s}$ and $\mathbf{r}$, respectively:
\begin{equation}
    \mathbf{\Sigma} = \mathbf{R}\mathbf{S}\mathbf{S}^T\mathbf{R}^T.
\end{equation}

During the rendering process, 3D Gaussians need to be projected onto the 2D image plane within a specific camera coordinate system. The covariance matrix $\mathbf{\Sigma}' \in \mathbb{R}^2$ in 2D space can be obtained from the view transformation matrix $\mathbf{W}$ and the Jacobian matrix $\mathbf{J}$ of the approximated projection transformation \cite{zwicker2001surface}:
\begin{equation}
    \mathbf{\Sigma}' = \mathbf{J}\mathbf{W}\mathbf{\Sigma} \mathbf{W}^T\mathbf{J}^T.
\end{equation}
The 3D Gaussians associated with each pixel can be sorted based on their respective depths. The pixel color $\mathbf{C}$ is computed by blending all $N$ Gaussians in the depth-sorted order:
\begin{equation}
    \mathbf{C} = \sum_{i=1}^N \mathbf{c_i} \alpha_i' \prod_{j=1}^{i-1} (1-\alpha_j'),
\end{equation}
where $\mathbf{c_i}$ is the color derived from the spherical harmonics coefficients of the $i$-th Gaussian with view direction, and $\mathbf{\alpha}_i'$ denotes the opacity obtained by the multiplication of the opacity of the $i$-th Gaussian with the covariance matrix $\mathbf{\Sigma}'$.

\section{Method}

The proposed ESGaussianFace framework is shown in Fig. \ref{method}. It mainly comprises three parts: a \textit{triplane-based 3D Gaussian generator} (Sec. \ref{mod1}), an \textit{audio-visual feature extraction and fusion module} (Sec. \ref{mod2}) and an \textit{ESGaussian deformation prediction module} (Sec. \ref{mod3}). 

The triplane-based 3D Gaussian generator decodes features generated by a triplane to obtain the Gaussian parameters of the canonical face. In the audio-visual feature extraction and fusion module , we extract both audio and emotion features from the input to guide the computation of the spatial attention. The ESGaussian deformation prediction module predicts the deformation of the Gaussian parameters under emotional and stylistic control, and uses a neural renderer to generate emotional and stylized talking head videos. Additionally, we introduce a \textit{multi-stage training strategy} (Sec. \ref{mod4}) designed to enable the model to effectively capture and predict deformations across a diverse range of emotions and styles.

\subsection{Triplane-based 3D Gaussian Generator}
\label{mod1}

In this section, we introduce the implementation specifics of the triplane-based 3D Gaussian generator. To achieve high-quality and 3D-consistent facial animation results, we employ 3DGS to obtain explicit 3D representations. We initialize $N$ sets of 3D Gaussians based on a 3DMM model of the standard human face. Subsequently, we encode the positional features of all 3D Gaussians using a multi-resolution tri-plane \cite{li2023efficient}. The tri-plane $T_c$ consists of three axis-aligned orthogonal feature planes. For any 3D position $\mathbf{\mu} \in \mathbb{R}^3$, we project it onto $T_c$ to obtain the corresponding feature vector $\mathbf{f}_p$:
\begin{equation}
    \mathbf{f}_p = (\mathbf{F}_{xy}, \mathbf{F}_{yz}, \mathbf{F}_{xz}) = T_c(\mathbf{\mu}).
\end{equation}
Each plane $\mathbf{F}$ has a resolution of $R \times R \times C$, where $R$ denotes the spatial resolution and $C$ represents the number of channels. $\mathbf{f}_p$ is then fed into a tri-plane decoder $D_c$, enabling the extraction of the canonical Gaussian parameters $\mathcal{G}_{can}$.

\subsection{Audio-visual Feature Extraction and Fusion Module}
\label{mod2}

This module performs the extraction of audio and video features, as well as the feature fusion via spatial attention. 

\subsubsection{Audio-visual Feature Extraction}

We aim to extract content features from the driving audio to guide the generation of lip movements in the human face. For an emotional audio segment $A^{1:T}$, we extract its Deep Speech features \cite{hannun2014deep} $\mathbf{a}^{1:T}$, which effectively capture the content of the audio. To capture contextual information, we extend $\mathbf{a}^t \in \mathbb{R}^{16 \times 29}$ by $l$ frames both forward and backward, resulting in $\mathbf{a}^{t-l:t+l}$ with a temporal length of $2l$. 

To imbue the generated results with different emotions, we use an emotional video (or an emotional image) as the emotion source. For emotion feature extraction, we employ a pretrained Deep3D \cite{deng2019accurate} model as the 3D facial extractor $E_d$. This model, based on 3DMM, performs 3D reconstruction and extracts 3D coefficients of 2D facial images. Here we use the expression coefficients $\mathbf{e} \in \mathbb{R}^{64}$ as the emotion feature. Following previous works, we extract the AU45 feature $y \in \mathbb{R}$ \cite{baltruvsaitis2016openface} from $I_{emo}$ to quantify the degree of blinking. Furthermore, we capture the extrinsic camera pose of the video and encode the viewpoint feature $\mathbf{v} \in \mathbb{R}^{12}$. 

\begin{figure}[t]
\centering
\includegraphics[width=\columnwidth]{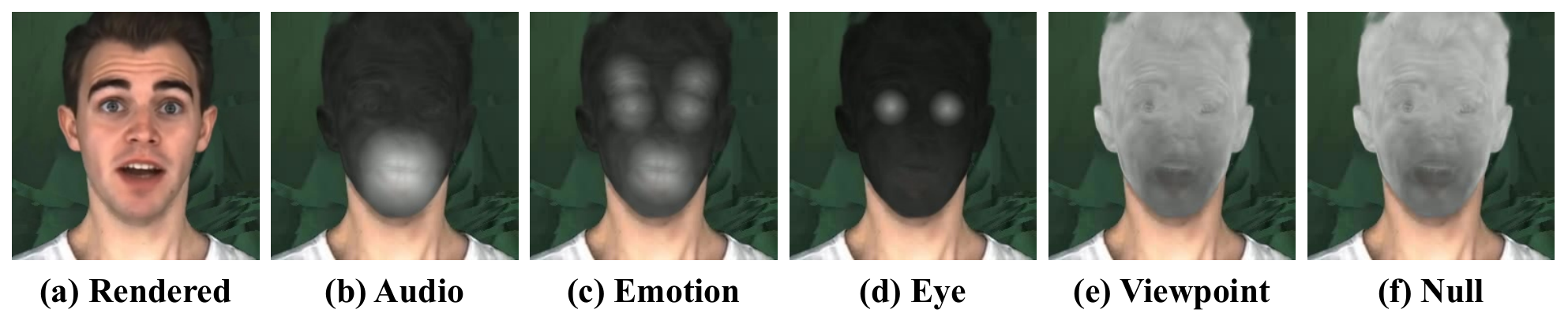} 
\caption{(a) Shows the rendered image, while the other panels display the attention score distributions for (b) audio content, (c) emotions, (d) eye blinks, (e) head orientations, and (f) temporal consistency.}
\label{att}
\end{figure}

\subsubsection{Emotion-audio-guided Spatial Attention} 

We aim for the audio and emotion features to influence different facial regions and guide the deformation of Gaussian points within them. For this purpose, we introduce an emotion-audio-guided spatial attention method based on spatial-audio attention \cite{cho2024gaussiantalker}. The spatial attention dynamically allocate weights to different regions in space, enabling the fusion of audio and emotion features while controlling the 3D Gaussian points. We integrates $B$ Emotion-audio-guided Spatial Attention Modules (ESAMs). Each ESAM is composed of two cross-attention layers and a Feed-forward (FF) layer. The first, known as the Audio-guided Attention (AGA) layer, forecasts the effect of audio on distinct facial regions. In contrast, the second layer, the Emotion-guided Attention (EGA) layer, estimates the influence of different emotions across various facial regions:
\begin{equation}
    \mathbf{z}_0 = \mathbf{f}_p,
\end{equation}
\begin{equation}
    \mathbf{z}_b^{\prime} = F_{ca}(\mathbf{z}_{b-1}, \mathbf{f}_{c}^{t}) + \mathbf{z}_{b-1}, \qquad b=1,...,B,
\end{equation}
\begin{equation}
    \mathbf{z}_b^{\prime\prime} = F_{eg}(\mathbf{z}_{b}^\prime, \mathbf{f}_{e}^{t}) + \mathbf{z}_{b}^\prime, \qquad b=1,...,B,
\end{equation}
\begin{equation}
    \mathbf{z}_b = F_{fl}(\mathbf{z}_b^{\prime\prime}) + \mathbf{z}_b^{\prime\prime}, \qquad b=1,...,B,
\end{equation}
where $\mathbf{f}_{a}^{t} = \{ \mathbf{a}^{t-l:t+l}, y, \mathbf{v}, \varnothing \}$, $\mathbf{f}_{e}^{t} = \{ \mathbf{e}, \varnothing \}$. $\varnothing$ is an empty vector, ensuring the consistency of global features across different frames. $F_{ag}$, $F_{eg}$ and $F_{fl}$ represent AGA, EGA and FF, respectively. 

We visualize the attention scores of different features across various facial regions, as shown in Fig. \ref{att}. It is evident that the audio features primarily affect areas around the mouth, while emotion features influence multiple regions that are more sensitive to emotional variations, such as the eyes, eyebrows and mouth. This significantly ensures the stability and accuracy of video generation in dynamic scenarios with changes of audio contents and emotional expressions. 

\subsection{ESGaussian Deformation Prediction Module}
\label{mod3}

To obtain the emotional and stylized ADFA results, we introduce the ESGaussian deformation prediction module. This module consists of two Gaussian parameter deformation predictors, $P_{emo}$ and $P_{sty}$. In addition to the spatially-aware feature $\mathbf{z}_B$, we incorporate the emotion feature $\mathbf{e}$ to further capture different emotions. We also discover that incorporating the positional embedding of 3D points yields more precise results. For this, we use an MLP network $E_{\mu}$ as encoder to obtain the encoded feature $\mathbf{p} \in \mathbb{R}^{64}$ of the 3D positions.

We input the feature $\mathbf{f}_{emo} = \{ \mathbf{z}_B, \mathbf{e}, \mathbf{p} \}$ into emotional deformation predictor $P_{emo}$ to obtain the Gaussian deformation $\Delta\mathcal{G}_{emo} = \{\Delta\mathbf{\mu}_{emo},\Delta\mathbf{s}_{emo},\Delta\mathbf{q}_{emo},\Delta\mathbf{sh}_{emo},\Delta\alpha_{emo} \}$. Adding $\Delta\mathcal{G}_{emo}$ to $\mathcal{G}_{can}$ yields $\mathcal{G}_{emo}$, which are input to the 3DGS neural renderer $\mathcal{R}$ to generate the image $\hat{I}_{emo}$ that reflects the emotional expression:
\begin{equation}
    \hat{I}_{emo} = \mathcal{R}(\mathcal{G}_{emo}) = \mathcal{R}(\mathcal{G}_{can} + P_{emo}(\mathbf{f}_{emo})).
\end{equation}

Regarding the stylized deformation predictor $F_{sty}$, it also utilizes the spatially-aware feature $\mathbf{z}_B$, the emotion feature $\mathbf{e}$ and the position feature $\mathbf{p}$ as inputs. Furthermore, we utilize a pretrained style encoder \cite{richardson2021encoding} and an MLP encoder jointly as style extractor $E_s$ to capture extrinsic style feature $\mathbf{s} \in \mathbb{R}^{128}$ of the style images. The feature $\mathbf{f}_{sty}=\{ \mathbf{z}_B, \mathbf{e}, \mathbf{p}, \mathbf{s} \}$ is then input into $P_{sty}$ to obtain the Gaussian deformation $\Delta\mathcal{G}_{sty}$. This deformation imparts the style of $I_s$ to the emotional 3D Gaussians, thereby rendering images $\hat{I}_{sty}$ that combine both the art style and emotional expression.

\begin{table*}[t]
        \caption{Quantitative comparisons of stylized emotional ADFA results with state-of-the-art methods. The top-performing results are highlighted in \textbf{bold}, while the second-best results are \underline{underlined}. For each method, We also provide its additional capabilities in generating emotional expression, art style, and multi-view images.}
        \label{tab:compar1}
	\centering
	\resizebox{\linewidth}{!}{
	\begin{tabular}{l|ccccccc|ccccccc|ccc}
		\toprule
		\multicolumn{1}{c}{\multirow{2}[3]{*}{\textbf{Method}}} & \multicolumn{7}{c}{\textbf{MEAD} \cite{wang2020mead}} & \multicolumn{7}{c}{\textbf{RAVDESS} \cite{livingstone2018ryerson}}  & \multicolumn{3}{c}{\textbf{Output}} \\ 
		\cmidrule(lr){2-8}  \cmidrule(lr){9-15} \cmidrule(lr){16-18}
        \multicolumn{1}{c}{} & \multicolumn{1}{c}{PSNR$\uparrow$} & \multicolumn{1}{c}{SSIM$\uparrow$} & \multicolumn{1}{c}{FID$\downarrow$} & \multicolumn{1}{c}{LPIPS$\downarrow$} & \multicolumn{1}{c}{Sync$\uparrow$} & \multicolumn{1}{c}
        {LMD$\downarrow$} & \multicolumn{1}{c}{$\mathrm{Acc_{emo}}$$\uparrow$} & \multicolumn{1}{c}
        {PSNR$\uparrow$} & \multicolumn{1}{c}{SSIM$\uparrow$} & \multicolumn{1}{c}{FID$\downarrow$} & \multicolumn{1}{c}{LPIPS$\downarrow$} & \multicolumn{1}{c}{Sync$\uparrow$} & \multicolumn{1}{c}
        {LMD$\downarrow$} & \multicolumn{1}{c}{$\mathrm{Acc_{emo}}$$\uparrow$} & \multicolumn{1}{c}
        {Emotion} & \multicolumn{1}{c}
        {Style} & \multicolumn{1}{c}
        {Multi-view}
		\\

		\midrule
		MakeItTalk \cite{zhou2020makelttalk} & 27.597 & 0.602 & 53.090 & 0.059 & 4.495 & 5.208 & 19.379 & 27.760 & 0.689 & 29.168 & 0.058 & 5.135 & 5.786 & 20.468 & \XSolidBrush & \XSolidBrush & \XSolidBrush   \\
        
        Wav2Lip \cite{prajwal2020lip} & 27.819 & 0.670 & 46.373 & 0.051 & \underline{5.851}  & 5.068 & 16.802 & 27.931 & 0.715 & 32.567 & 0.057 & \textbf{6.169} & 5.623 & 18.027 & \XSolidBrush & \XSolidBrush & \XSolidBrush \\
        
        Audio2Head \cite{wang2021audio2head} & 26.529 & 0.623 & 63.287 & 0.070 & 3.342 & 6.681 & 19.915 & 25.639 & 0.583 & 69.613 & 0.069 & 2.837 & 7.243 & 19.322 & \XSolidBrush & \XSolidBrush & \XSolidBrush   \\	
        
        MEAD \cite{wang2020mead} & 28.081 & 0.712 & \underline{21.273} & 0.039 & 5.784 & 3.564 & 64.920 & -- & -- & -- & -- & -- & -- & -- & \Checkmark & \XSolidBrush & \XSolidBrush \\
        
		EAMM \cite{ji2022eamm} & 26.807 & 0.681 & 58.334 & 0.063 & 3.457 & 4.578 & 25.415 & 26.071 & 0.643 & 50.811 & 0.068 & 4.106 & 5.746 & 27.133 & \Checkmark & \XSolidBrush & \XSolidBrush \\

        EAT \cite{gan2023efficient} & 27.650 & 0.690 & 35.094 & 0.049 & 5.277 & 4.360 & 58.485 & 26.134 & 0.613 & 40.736 & 0.067 & 3.662 & 6.459 & 53.869 & \Checkmark & \XSolidBrush & \XSolidBrush \\

        DreamTalk \cite{ma2023dreamtalk} & 27.801 & 0.732 & 34.239 & 0.051 & 5.560 & 4.693 & 61.810 & 26.193 & 0.650 & 38.826 & 0.064 & 5.076 & 6.118 & 57.090 & \Checkmark & \XSolidBrush & \XSolidBrush \\

        EDTalk \cite{tan2024edtalk} & 27.938 & 0.760 & 34.497 & 0.048 & 5.436 & 4.467 & 62.288 & 26.466 & 0.667 & 36.221 & 0.060 & 4.720 & 5.854 & \underline{59.315} & \Checkmark & \XSolidBrush & \XSolidBrush \\
        
		Style$^2$Talker \cite{tan2024style2talker} & \underline{29.455} & 0.795 & 25.282 & \underline{0.035} & 5.294 & \underline{3.007} & \underline{70.273} & 26.906 & 0.647 & 41.649 & 0.061 & 3.410 & 6.186 & 49.544 & \Checkmark & \Checkmark & \XSolidBrush  \\
		
		AD-NeRF \cite{guo2021ad} & 25.282 & 0.549 & 84.506 & 0.086 & 1.298 & -- & 10.245 & 25.021 & 0.508 & 86.448 & 0.082 & 1.132 & -- & 10.714 & \XSolidBrush & \XSolidBrush & \Checkmark  \\

        SyncTalk \cite{peng2024synctalk} & 28.120 & 0.803 & 32.959 & 0.040 & 4.770 & 5.071 & 48.245 & 27.706 & 0.697 & 31.793 & 0.053 & 3.857 & 5.646 & 47.680 & \XSolidBrush & \XSolidBrush & \Checkmark  \\

		GaussianTalker \cite{cho2024gaussiantalker} & 28.911 & \underline{0.818} & 22.290 & 0.038 & 5.202 & 4.981 & 52.621 & \underline{28.516} & \underline{0.747} & \underline{25.260} & \underline{0.044} & 4.354 & \underline{5.329} & 51.319 & \XSolidBrush & \XSolidBrush & \Checkmark \\ 
        
		\midrule
		\textbf{ESGaussianFace} & \textbf{31.873} & \textbf{0.901} & \textbf{16.933} & \textbf{0.028} & \textbf{6.216} & \textbf{2.832} & \textbf{75.944} &  \textbf{30.772} & \textbf{0.833} & \textbf{21.796} & \textbf{0.034} & \underline{5.757} & \textbf{3.145} & \textbf{71.124} & \Checkmark & \Checkmark & \Checkmark  \\
		\bottomrule
	
	\end{tabular}%
	}
\end{table*}

\subsection{Multi-stage Training Strategy}
\label{mod4}

During training, the Gaussian deformation predictor often fails to directly learn the deformations from a neutral face to faces with lip movements and different emotions. Directly training the model leads to weak emotional expressions, loss of facial features, or even distortion. To address this, we propose a multi-stage training strategy to obtain more accurate results. 

\subsubsection{Neutral Stage with Lip Movement}

First, we train the triplane-based 3D Gaussian generator $M_G$ and the emotional deformation prediction module $M_P^{emo}$ ($E_\mu$ and $P_{emo}$). The goal of this stage is to enable the $M_G$ to accurately generate the Gaussian parameters $\mathcal{G}_{can}$ of a canonical face, while allowing $M_P^{emo}$ to learn the lip movement of neutral emotion. We use the $t$-th frame of the talking head video $I_{neu}^{1:T}$, which depicts neutral emotion, for training supervision. We employ a loss function that comprises six distinct components: 
\begin{equation}
\begin{split}
   L &= \lambda_{1} L_{rgb} + \lambda_{2} L_{per} + \lambda_{3} L_{ssim} \\
   &+ \lambda_{4} L_{lip} +\lambda_{5} L_{ld} + \lambda_{6} L_{smo},
\end{split}
\end{equation}
where $L_{rgb}$, $L_{per}$, and $L_{ssim}$ represent the L1 color loss, perceptual loss \cite{johnson2016perceptual}, and SSIM loss \cite{wang2004image}, respectively; $\lambda_i$ denotes the weighting factor of the loss function. To ensure consistency of lip movements with the audio, we employ L1 loss to the mouth region to accurately align the lip movements: 
\begin{equation}
    L_{lip} = \| m_{neu} \cdot I_{neu} - \hat{m}_{neu} \cdot \hat{I}_{neu} \|_1,
\end{equation}
where $m_{neu}$ denotes the mask of the mouth region extracted by a face parsing model \cite{lee2020maskgan}. Furthermore, we utilize the facial landmarks to ensure the accuracy of facial structure and emotions. This leads to the landmark distance loss $L_{ld}$:
\begin{equation}
    L_{ld} = \| \mathbf{w}(F_{ld}(I_{neu}) - F_{ld}(\hat{I}_{neu})) \|_2^2,
\end{equation}
$F_{ld}$ denotes the pretrained landmark detector, and $w_i \in \mathbf{w}$ is the weight for the $i$-th landmark. Moreover, we add a smooth loss $L_{smo}$ to eliminate temporal jitter in generated videos: 
\begin{equation}
    L_{smo} = \| \Delta\mathcal{G}_{neu}^{t} - 0.5 \times (\Delta\mathcal{G}_{neu}^{t-1} + \Delta\mathcal{G}_{neu}^{t+1}) \|_2^2.
\end{equation}

In this stage, we select talking head videos exhibiting neutral emotion of a specific actor from the emotional dataset. The objective of this stage is to enable $M_G$ to learn the person's explicit 3D representation, while allowing $M_F$ and $M_P^{emo}$ to learn lip movements under varying audio conditions.

\subsubsection{Emotion Stage with Emotional Deformation}

Building on the pretraining of the model, the goal of this stage is to train the entire network (excluding the stylization component). Here, we fix the weights of $M_G$. After pretraining, $M_P^{emo}$ is capable of predicting lip movements of neutral emotion. At this point, the model only needs to predict the emotional deformation of the Gaussian parameters with the same lip movement. We supervise the model's generated $\hat{I}_{emo}$ using the same loss function as in the neutral stage for training.

Since the model has already learned the 3D facial representation of the current actor, we fix the parameters of $M_G$ in emotion stage. We select talking head videos of the same actor and the same audio but with different emotions for training. As $M_F$ and $M_P^{emo}$ are already capable of predicting lip movements corresponding to a given audio, this stage simply continues training these two modules based on the parameters obtained in the neutral stage, enabling them to learn the overall facial emotional deformation under consistent lip movements.

\subsubsection{Stylization Stage with 3DGS Style Transfer}

To further train the stylization functionality, we introduce a third training stage. We employ the VToonify \cite{yang2022Vtoonify} model on the emotional video $I_{emo}^{1:T}$ used in emotion stage and the style image $I_s$ to achieve stylization of emotional facial video, resulting in $I_{sty}^{1:T}$. In this stage, $M_P^{sty}$ ($E_s$ and $P_{sty}$) aims to learn the stylized deformation of the 3D Gaussian parameters guided by the style features $\mathbf{s}$. In addition to the loss functions from $L$, we introduce a extrinsic style loss $L_{sty}$ to ensure consistency between the generated results and the style of $I_s$:
\begin{equation}
    L_{exs} = \| E_s(I_{sty}) - E_s(\hat{I}_{sty}) \|_1.
\end{equation}

After completing the multi-stage training, we obtain a person-specific 3D Gaussian model capable of controlling arbitrary emotions and styles. During inference, since $M_G$ has already learned the 3D Gaussian representation of the target person, no additional input image of the individual is required. We only need to provide the target driving audio, an image representing the target emotion (emotion source), and an image with the desired art style (style source). ESGaussianFace can then generate a 3D talking head with \textit{accurate lip synchronization}, \textit{realistic emotional expression}, and \textit{faithful stylization}.

\begin{figure*}[t]
\centering
\includegraphics[width=0.986\textwidth]{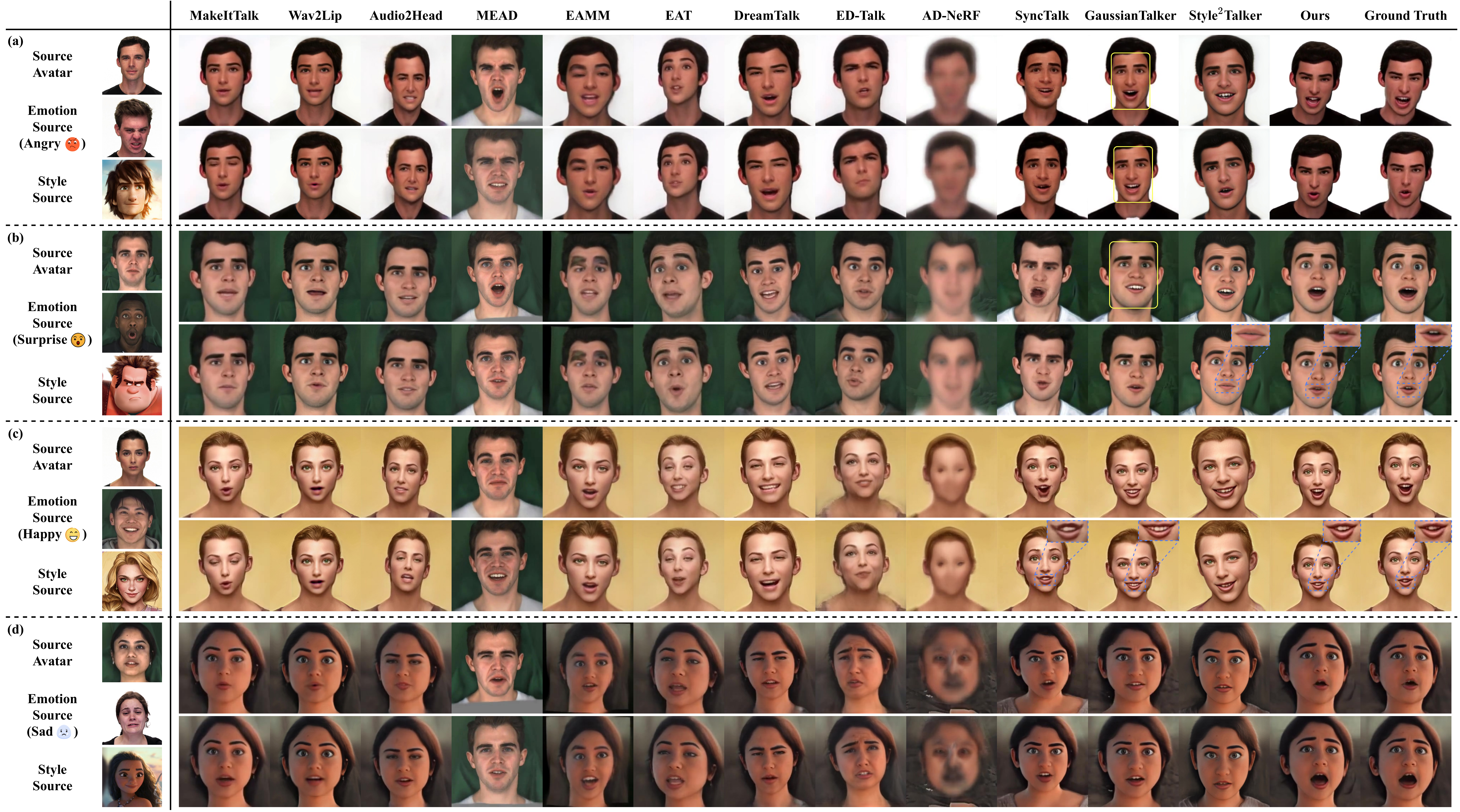} 
\caption{Qualitative comparisons of stylized emotional ADFA results with state-of-the-art methods. Experiments (a) and (b) present results on the RAVDESS and MEAD datasets, respectively, where the source avatar and emotion source are derived from the same dataset in each experiment. Experiments (c) and (d) show results with the emotion source from different datasets.}
\label{compar}
\end{figure*} 

\section{Experiments}

\subsection{Experimental Settings}
\subsubsection{Datasets}
In our experiments, we use the MEAD \cite{wang2020mead} and RAVDESS \cite{livingstone2018ryerson} datasets for training. MEAD is a high-quality publicly available dataset that contains audio-visual data of 8 emotions performed by 60 actors. To demonstrate the generalizability of our model across different datasets, we also incorporate the RAVDESS dataset, which consists of audio-visual data representing 8 emotions from 24 actors. Additionally, we use various art datasets \cite{pinkney2020resolution} to obtain art style references. For further details on the datasets and experimental parameters, please consult the supplementary materials.

\subsubsection{Comparison Setting} 
To perform a comprehensive comparison of talking head videos with both emotional expressions and art styles, we follow the experimental approach of Style$^2$Talker \cite{tan2024style2talker}. We input the source facial image and the style image into VToonify \cite{yang2022Vtoonify} to generate a stylized facial image. This image is then processed using several state-of-the-art ADFA methods, achieving results comparable to our method. We select three traditional ADFA methods: MakeItTalk \cite{zhou2020makelttalk}, Wav2Lip \cite{prajwal2020lip}, and Audio2Head \cite{wang2021audio2head}. In the category of ADFA methods based on NeRF or 3DGS, we choose AD-NeRF \cite{guo2021ad}, SyncTalk \cite{peng2024synctalk} GaussianTalker \cite {cho2024gaussiantalker}. For emotional ADFA methods, we include MEAD \cite{wang2020mead}, EAMM \cite{ji2022eamm}, EAT \cite{gan2023efficient}, DreamTalk \cite{ma2023dreamtalk} and EDTalk \cite{tan2024edtalk} in our comparisons. 

We employ PSNR, SSIM \cite{wang2004image}, FID \cite{heusel2017gans}, and LPIPS \cite{zhang2018unreasonable} to evaluate the quality of the generated images and their similarity to real images. To compare the consistency of lip movements with the audio, we use the confidence score of SyncNet \cite{chung2017out} (Sync). Additionally, we measure the accuuracy of expressions and poses by the average distance between landmarks \cite{cheng2024stylizedfacepoint} of the generated and real faces (LMD), and further assess emotional accuracy using $\mathrm{Acc_{emo}}$ \cite{meng2019frame}.

\subsection{Experimental Results}

\subsubsection{Quantitative Results}
The quantitative comparison of the stylized emotional ADFA results between our method and state-of-the-art methods is given in Tab. \ref{tab:compar1}. As observed, we are the only method that supports generation of emotional expressions, art styles, and multi-view rendering. It consistently outperforms others on most evaluation metrics, ranking second only to Wav2Lip in Sync. This is attributed to Wav2Lip's use of SyncNet as discriminator during training. Notably, the lowest LMD demonstrates that our method is the most accurate in representing lip movements and emotional expressions. 

\subsubsection{Qualitative Results}
Fig. \ref{compar} presents a qualitative comparisons of the results. Experiments (a) and (b) present test results on the RAVDESS and MEAD datasets, respectively, where the source avatar and emotion source are derived from the same dataset in each experiment. Experiments (c) and (d) show results with the emotion source selected from different datasets. For stylization, experiments (a) and (b) show the results without color transfer, while experiments (c) and (d) include color transfer. As observed, while GaussianTalker excels at restoring facial poses and details, it tend to confuse different emotions (circled in \textit{yellow} boxes). Compared to other methods, our results demonstrate significantly improved accuracy in lip movements (indicated by \textit{blue} boxes). Furthermore, our method supports multi-view video generation, showcasing its versatility and effectiveness. More experimental results and an in-depth analysis are provided in the supplementary material.

\begin{table*}[t]
  \caption{We use FPS to measure the efficiency of reenactment. The top-performing results are highlighted in \textbf{bold}, while the second-best results are \underline{underlined}.}
  \label{tab:fps}
  \centering
  \resizebox{\linewidth}{!}{
  \begin{tabular}{c|ccccccccccccc}
    \toprule
    \textbf{Method} & MakeItTalk & Wav2Lip & Audio2Head & MEAD & EAMM & EAT & DreamTalk & EDTalk & Style$^2$Talker & AD-NeRF & SyncTalk & GaussianTalker & ESGaussianFace\\
    \midrule
    \textbf{FPS} & 14.290 & 15.243 & 13.817 & 5.715 & 8.351 & 15.360 & 7.832 & 16.878 & 14.905 & 0.112 & 1.030 & \textbf{76.802} & \underline{69.624}\\
  \bottomrule
\end{tabular}
}
\end{table*}

\subsubsection{Inference Speed} 

We evaluate the efficiency using FPS. Our method achieves an FPS of 69, enabling real-time rendering and generation. As shown in Tab. \ref{tab:fps}, our method outperforms most state-of-the-art methods in terms of efficiency, ranks second only to GaussianTalker. This is attributed to the incorporation of additional emotion and stylization modules. Nevertheless, the lightweight predictor in our model enables flexible control over arbitrary emotions and styles, resulting in only a minimal FPS difference compared to GaussianTalker.

\begin{figure}[t]
\centering
\includegraphics[width=\columnwidth]{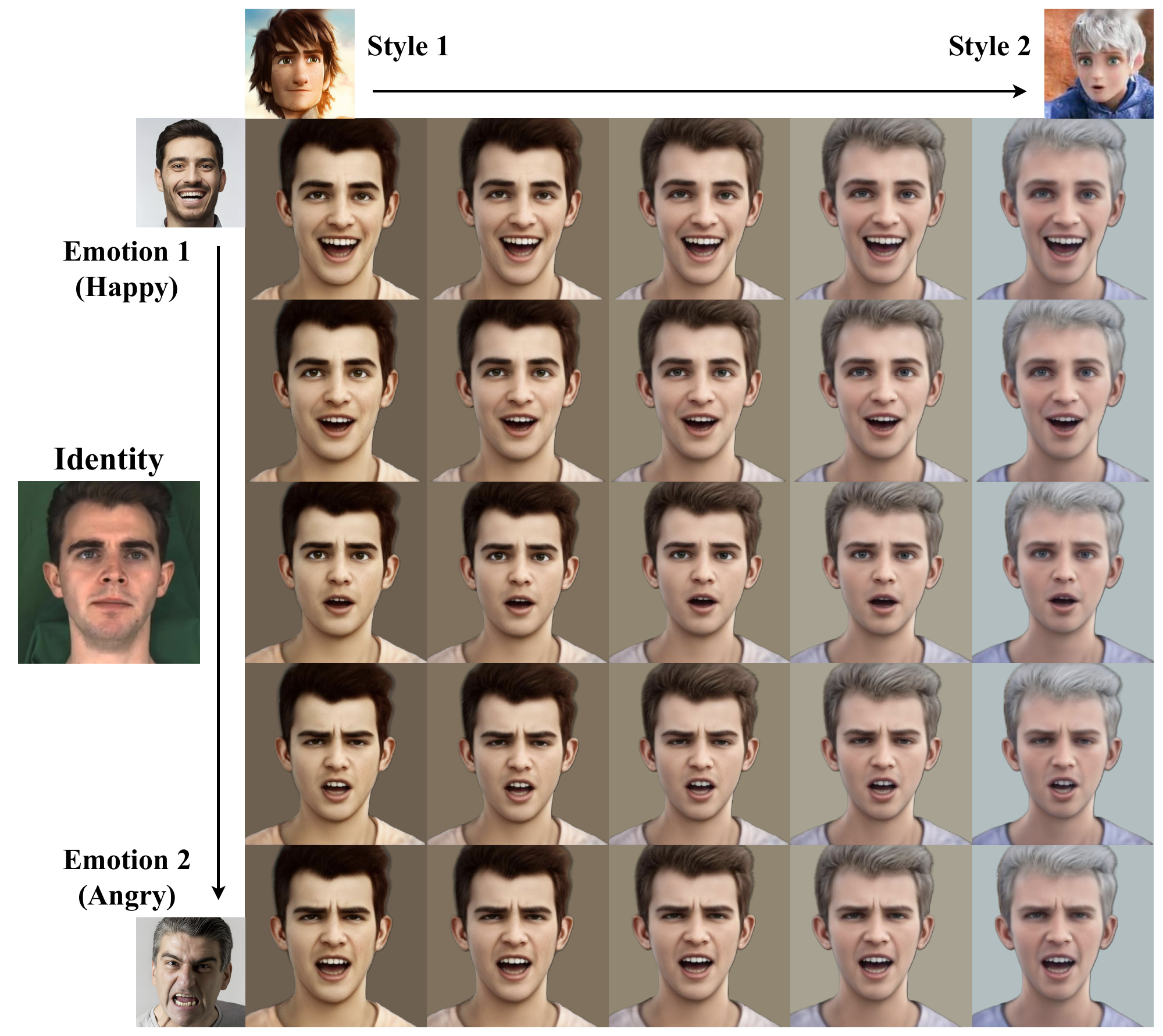} 
\caption{Emotion and style manipulation results. We set the interpolation weight to 1, 0.75, 0.5, 0.25, and 0, respectively.}
\label{inter}
\end{figure}

\subsubsection{Emotion and Style Manipulation}

We can achieve emotion and style manipulation through linear interpolation: 
\begin{equation}
    \mathbf{f} = \alpha \mathbf{f}_1 + (1-\alpha) \mathbf{f}_2.
\end{equation}
Here, $\mathbf{f}$ denotes the emotion or style feature, and $\alpha$ represents the interpolation weight. For emotion manipulation, we use the emotion feature $\mathbf{e}$ encoded by 3D facial extractor $E_d$. For style manipulation, we use the style feature $\mathbf{s}$ extracted from the style source $I_s$. The results of emotion and style manipulation under different interpolation weights are shown in Fig. \ref{inter}. We set the value of $\alpha$ to 1, 0.75, 0.5, 0.25, and 0, respectively. As shown in the figure, our method successfully achieves smooth and continuous transitions in both emotion and style domains by interpolating the corresponding features.

\subsubsection{Multi-view Rendering Results}

\begin{figure}[t]
\centering
\includegraphics[width=\columnwidth]{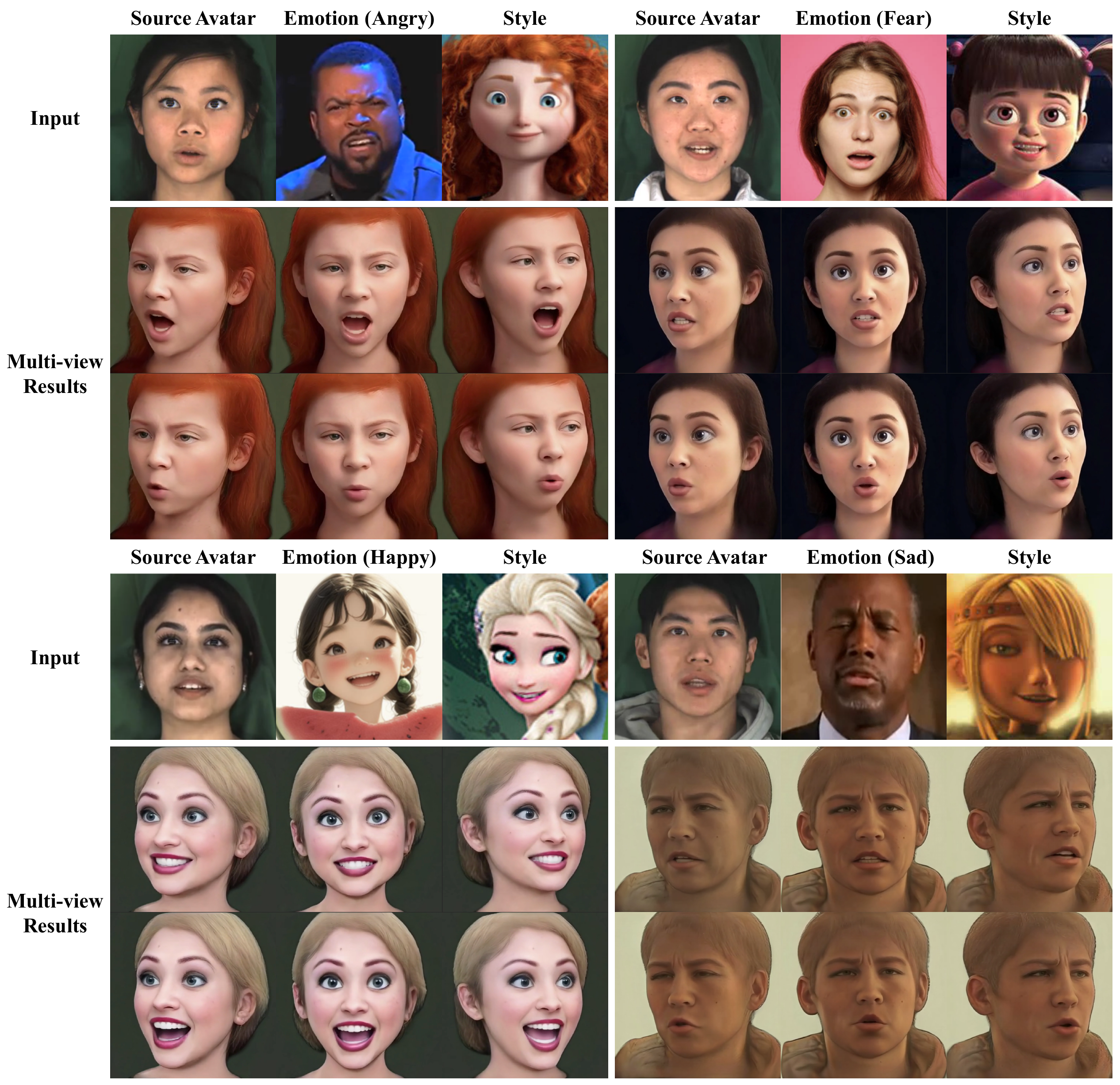} 
\caption{Multi-view rendering results. We select \textit{in-the-wild} emotional images as emotion sources.}
\label{multiview}
\end{figure}

Our method enables multi-view rendering based on 3D Gaussian Splatting. We present rendering results from various viewpoints, as shown in Fig. \ref{multiview}. We select in-the-wild images representing angry, surprised, happy and sad emotions as emotion sources. The results demonstrate that our method can accurately control both the target emotion and style, while maintaining strong 3D consistency across different rotation angles.

\subsubsection{Generalization on Other Datasets}

\begin{figure}[t]
\centering
\includegraphics[width=\columnwidth]{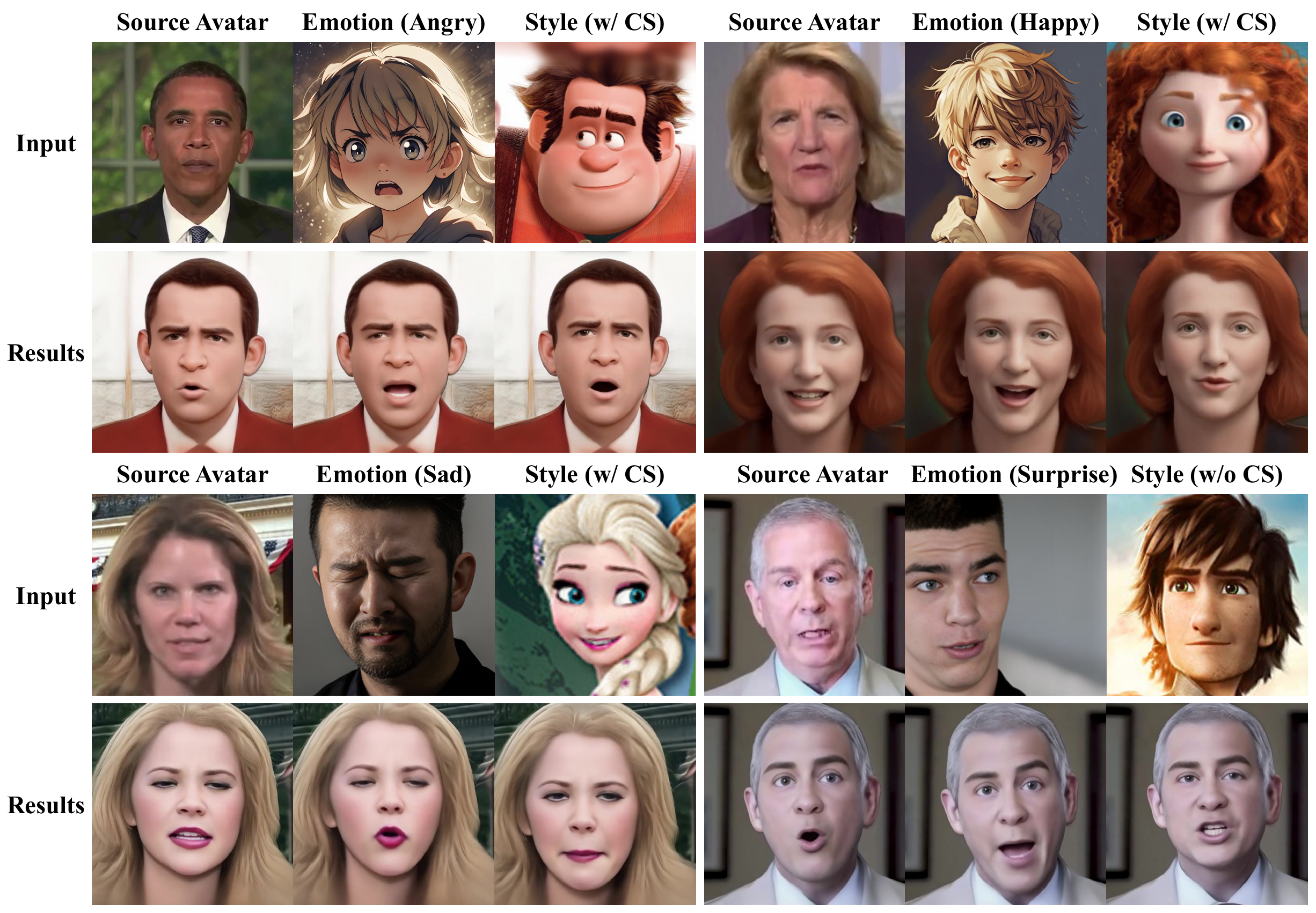} 
\caption{Our method demonstrates strong generalization capabilities to videos from other datasets \cite{guo2021ad, zhang2021flow}.}
\label{inthewild}
\end{figure}

Most videos outside of the emotional datasets \cite{wang2020mead, livingstone2018ryerson} exhibit neutral emotions, making it challenging to find emotional videos that meet the requirements of our training strategy. To demonstrate the generalization capability of our method, we utilize EmoStyle \cite{azari2024emostyle} to edit the emotion of each frame of any in-the-wild talking head video. These emotion-edited videos are then used to train our method. In Fig. \ref{inthewild}, we present inference results on videos beyond MEAD and RAVDESS datasets. These results demonstrate the strong generalization ability of our method.

\subsubsection{User Study}
\begin{figure}[t]
\centering
\includegraphics[width=0.97\columnwidth]{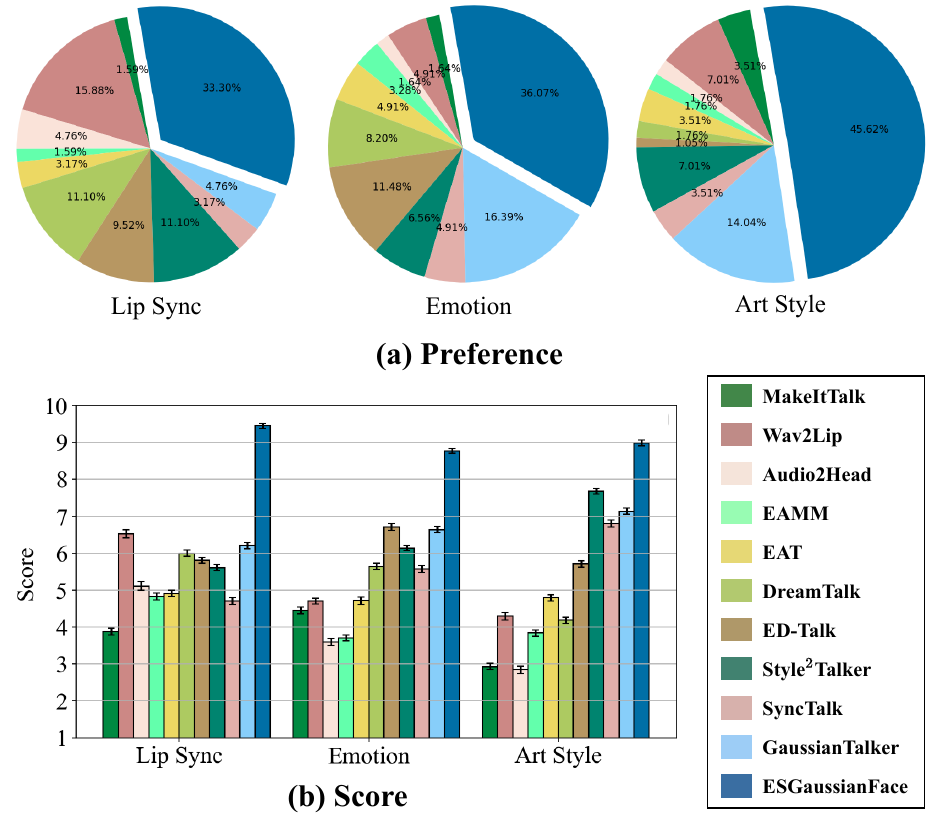} 
\caption{User Study Results. (a) The most preferred method of each participant; (b) The score ranges from 1 to 11, and error bars imply the standard deviations.}
\label{use_res}
\end{figure}

We conduct a user study to compare the performance with state-of-the-art methods. We recruit 24 participants (12 males and 12 females) and each is presented with 7 videos generated by 11 methods. Since MEAD can only generate driving results for a specific avatar and the outputs of AD-NeRF are blurry, we do not include them in the user study. Participants are asked to rate the accuracy of lip movement, emotional expressions, and art style in the generated videos, using a scale from 1 to 11 for different methods. We then identify the top-performing method for each participant and illustrate their preferences using a pie chart, which is included in Fig. \ref{use_res} (a). Additionally, we calculate the average scores for each method. The results for the three metrics are presented in Fig. \ref{use_res} (b). Our method demonstrate superior performance in terms of lip movements, emotion and art style. 

\subsubsection{Ablation Study}

\begin{table}
  \caption{Quantitative Results for ablation study.}
  \label{tab:ablation}
  \centering
  \resizebox{\columnwidth}{!}{
  \begin{tabular}{@{}c|ccccccc@{}}
    \toprule
    \textbf{Method} & PSNR $\uparrow$ & SSIM $\uparrow$ & FID $\downarrow$ & LPIPS $\downarrow$ & Sync $\uparrow$ & LMD $\downarrow$ & $\mathrm{Acc_{emo}}$ $\uparrow$ \\
    \midrule

    w/o ESAM & 28.679 & 0.769 & 25.403 & 0.040 & 4.876 & 4.982 & 29.807\\	
    					
    w/o MTS & 28.050 & 0.748 & 30.672 & 0.046 & 2.154 & 4.781 & 36.263\\	

    w/o $\mathbf{p}$ & 28.780 & 0.813 & 24.012 & 0.033 & 4.169 & 4.026 & 63.558\\	

    w/o $\mathbf{e}$ & 29.455 & 0.814 & 23.124 & 0.036 & 5.061 & 4.816 & 27.921 \\	

    w/o $L_{lip}$ & 30.025 & 0.878 & 18.539 & 0.029 & 3.064 & 3.524 & 66.542 \\	

    w/o $L_{ld}$  & 30.842 & 0.823 & 22.094 & 0.033 & 4.440 & 4.776 & 53.540\\	

    w/o $L_{exs}$ & 29.189 & 0.790 & 26.303 & 0.038 & 5.435 & 3.022 & 71.251\\	
    \midrule			
    \textbf{Full Model} & $\textbf{31.873}$ & $\textbf{0.901}$& $\textbf{16.933}$ & $\textbf{0.028}$ & $\textbf{6.216}$ & $\textbf{2.832}$ & $\textbf{75.944}$\\
    \bottomrule
  \end{tabular}
  }
\end{table}

\begin{figure}[t]
\centering
\includegraphics[width=0.98\columnwidth]{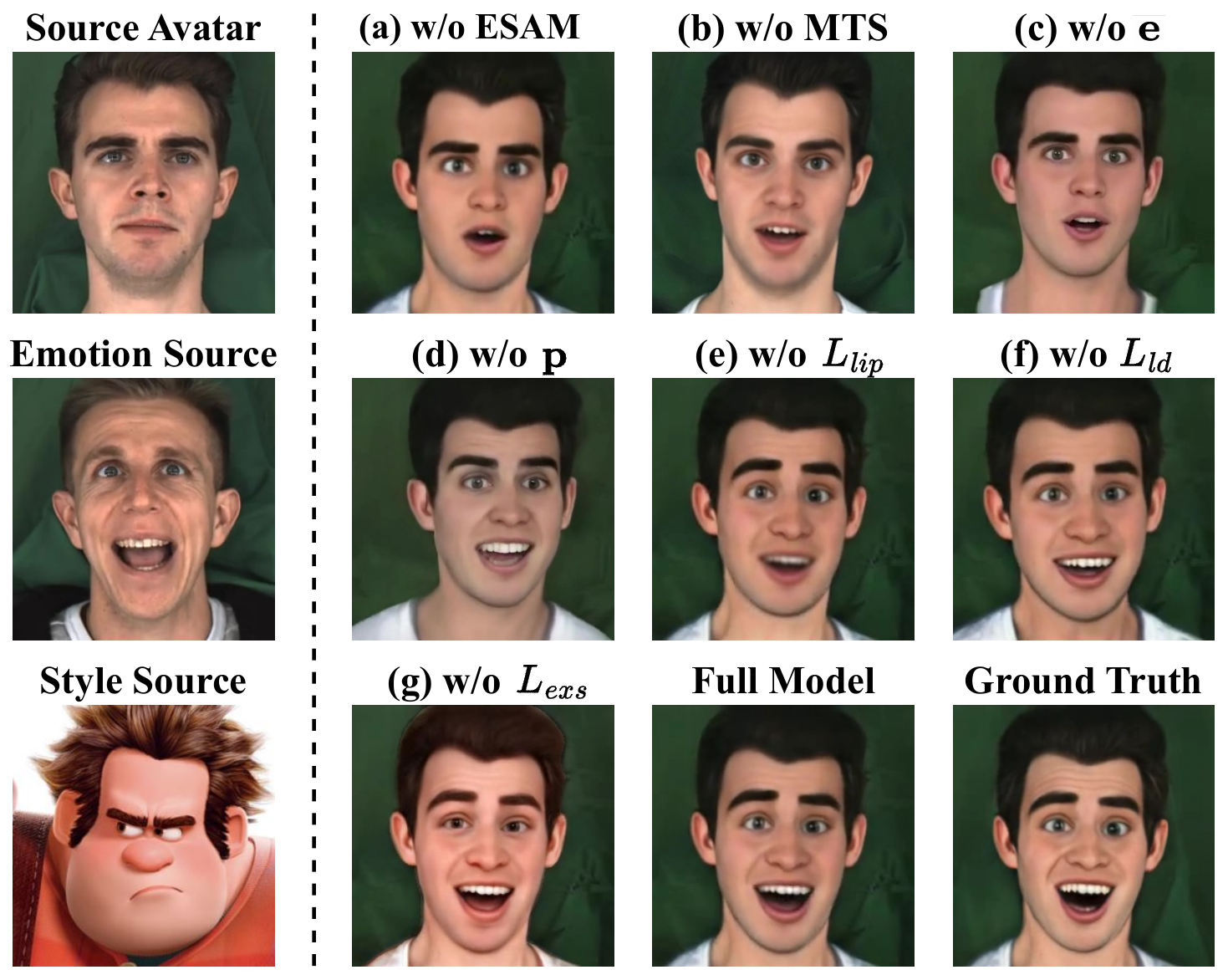} 
\caption{Qualitative ablation study results. The figure presents an example of ADFA results depicting a happy emotion.}
\label{fig:ablation}
\end{figure} 

To better demonstrate the effectiveness of our method's components, we conduct ablation experiments. The quantitative results are presented in Tab. \ref{tab:ablation}, while the qualitative results are shown in Fig. \ref{fig:ablation}. 

We design 7 variants in total: \textbf{(a)} First, we examine the contribution of the emotion-audio-guided spatial attention module introduced in Sec. \ref{mod2} (w/o ESAM). We replace our proposed ESAM module with the spatial-audio attention module employed in GaussianTalker. As shown in Fig. \ref{fig:ablation} (a), the model’s generated results, without ESAM, exhibit inaccurate emotional expressions. Furthermore, the lower $\mathrm{Acc_{emo}}$ observed in the quantitative results also indicates a decline in the accuracy of emotion reconstruction in the absence of ESAM module. The result highlights the crucial role of this module in controlling the character's emotional expression. \textbf{(b)} Next, we investigate the necessity of the multi-stage training strategy proposed in Sec. \ref{mod4} (w/o MTS). We bypass the first two stages and directly train the full model using stylized emotional videos as supervision. As seen in the results depicted in Fig. \ref{fig:ablation} (b), the output only shows slight color changes and does not achieve the desired stylization effect. Additionally, the emotional variation in the generated results is minimal, and the accuracy of lip movements shows a slight decline. These experimental findings suggest that directly training the full model for such a complex task is inadequate. A multi-stage approach, where lip movements, emotional expressions, and style features are learned incrementally, is essential. 

We also evaluate the necessity of two features, $\mathbf{e}$ and $\mathbf{p}$, used in the ESGaussian deformation prediction module. \textbf{(c)} Although the expression feature $\mathbf{e}$ is incorporated during the extraction of $\mathbf{z}_B$, it remains essential during the Gaussian parameter prediction process. In ESAM, $\mathbf{e}$ primarily assigns spatial attention weights across different facial regions. In contrast, $\mathbf{e}$ provides emotion features to guide the emotion-aware deformation of Gaussian parameters in the deformation prediction module. Fig. \ref{fig:ablation} (c), LMD and $\mathrm{Acc_{emo}}$ in Tab. \ref{tab:ablation} show that omitting $\mathbf{e}$ during the prediction process leads to inaccurate emotions. \textbf{(d)} 3D Gaussian points at different locations undergo varying degrees of deformation. Positional embedding feature $\mathbf{p}$ provides spatial location priors, improving the network’s understanding of emotion’s impact on different facial regions. The visualized results of Fig. \ref{fig:ablation} (d) and the PSNR values in Tab. \ref{tab:ablation} show that removing $\mathbf{p}$ leads to a noticeable decline in both image quality and accuracy.

Furthermore, We analyze the loss functions proposed in our method: \textbf{(e)} w/o $L_{lip}$: the absence of this loss primarily leads to a decrease in lip synchronization; \textbf{(f)} w/o $L_{ld}$: the lack of landmark constraints results in declines in various metrics of the results; \textbf{(g)} w/o $L_{exs}$: the absence of style loss does not significantly impact lip movement and landmark positions but greatly reduces the accuracy of the generated images. Both quantitative and qualitative results demonstrate that our proposed loss functions are essential for training our model.

\section{Limitation}

Despite the success of our work, we also recognize some limitations. \textbf{(a)} First, since the MEAD dataset contains only 3 intensity levels for each emotion, our method tends to produce similar results when processing emotion sources with subtle differences. \textbf{(b)} Our method uses stylized videos processed by VToonify as ground truth. However, more advanced stylization methods based on diffusion models have recently emerged. \textbf{(c)} Our method can only train a 3D Gaussian field for a specific character. To drive the talking head videos of different characters, separate 3D Gaussian models need to be trained for each. This is a common challenge faced by current ADFA methods based on NeRF and 3D Gaussians. In future work, we will focus on addressing these issues. 

\section{Conclusion}

In this paper, we introduce ESGaussianFace, a novel framework for generating high-quality talking head videos that incorporate both emotional expressions and style features. Leveraging 3D Gaussian Splatting, our method ensures high efficiency, 3D consistency, and multi-view rendering capabilities. We design an emotion-audio-guided spatial attention module that captures the influence of audio and emotion on the position of Gaussian points. The features output by this module, together with the encoded expression features and the embedding of 3D point positions, ensure the precision of the generated facial structures. Furthermore, to tackle this complex task, we propose a multi-stage training strategy, where the model learns the lip movements, emotional deformations, and style features in three stages. Both qualitative and quantitative results on multiple datasets demonstrate the superior performance of our approach over existing state-of-the-art methods.

\section*{Acknowledgments}
This work was completed during a visit to MSRA under the StarTrack program, hosted by Jiaolong Yang and Xin Tong. This work was supported in part by National Natural Science Foundation of China (NSFC, No. 62472285 and No. 62102255). 


\bibliography{IEEEexample}
\bibliographystyle{IEEEtran}

\section{Biography Section}

\begin{IEEEbiography}
[{\includegraphics[width=1in,height=1.25in,clip,keepaspectratio]{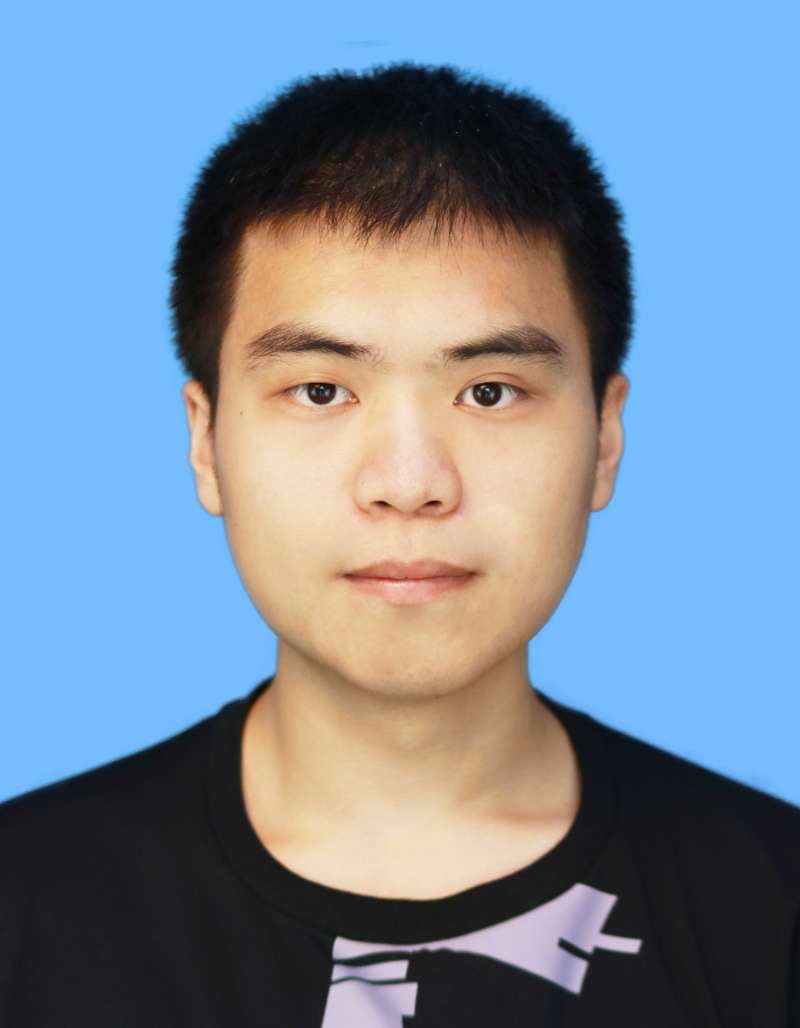}}]{Chuhang Ma}
is a member of Character Lab, Shanghai Jiao Tong University, Shanghai, China. He received the B.E. degree in Artificial Intelligence from Shanghai Jiao Tong University. He is currently pursuing the Ph.D. degree in Computer Science and Engineering at Shanghai Jiao Tong University. His research interest includes computer vision and 3D facial animation. 
\end{IEEEbiography}


\begin{IEEEbiography}
[{\includegraphics[width=1in,height=1.25in,clip,keepaspectratio]{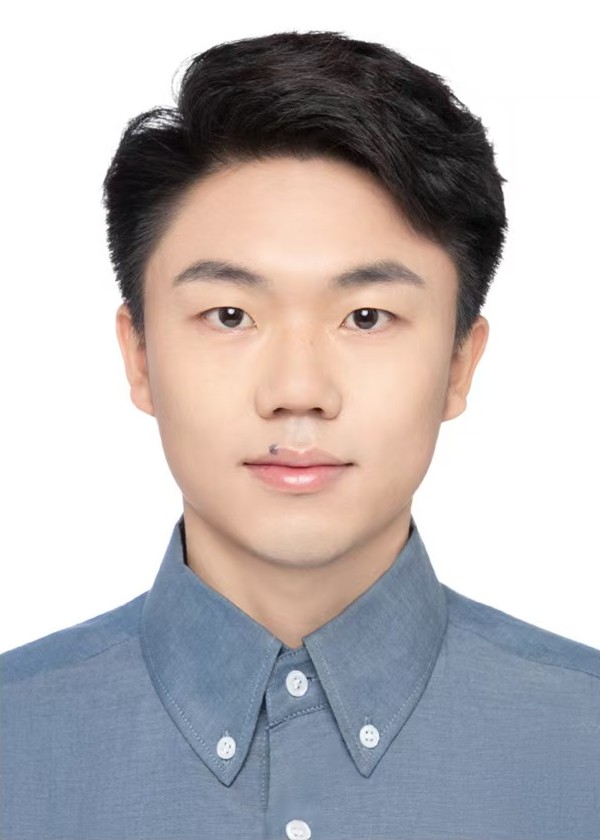}}]{Shuai Tan}
is a member of Character Lab, Shanghai Jiao Tong University, Shanghai, China. He received the B.S. degree in software engineering from Sichuan University. He is currently pursuing the Ph.D. degree in Computer Science and Engineering at Shanghai Jiao Tong University. His research interest includes computer vision and multi-modal learning.
\end{IEEEbiography}


\begin{IEEEbiography}
[{\includegraphics[width=1in,height=1.25in,clip,keepaspectratio]{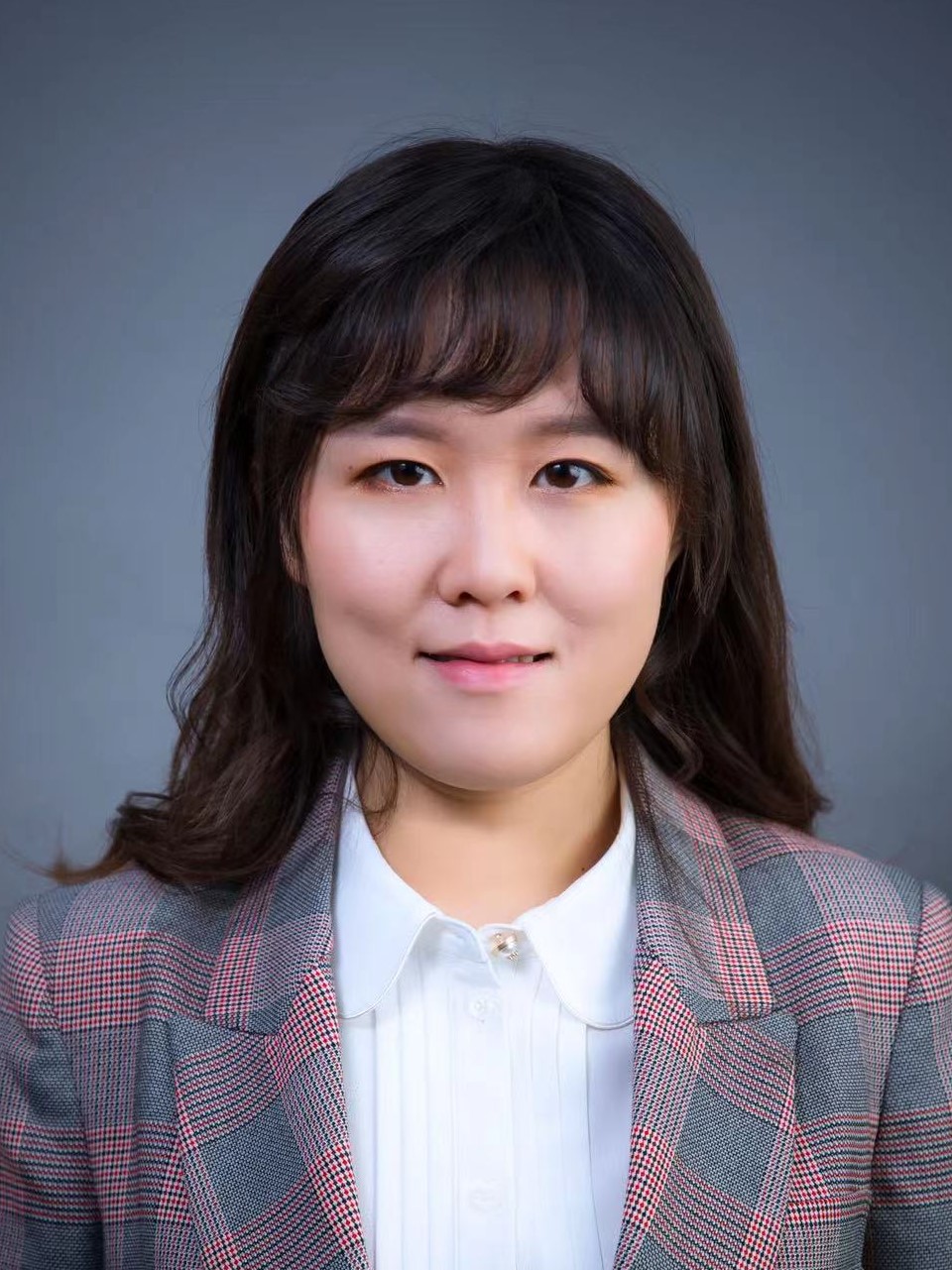}}]{Ye Pan}
is currently an Associate Professor with Shanghai Jiao Tong University. Her research interests include AR/VR, avatars/characters, 3D animations, HCI, and computer graphics. Previously, she was an Associate Research Scientist in AR/VR at Disney Research Los Angeles. She received the B.Sc. degree in communication and information engineering from Purdue/UESTC in 2010 and the Ph.D. degree in computer graphics from the University College London (UCL) in 2015. She has served as Associate Editor of the International Journal of Human Computer Studies, and a regular member of IEEE virtual reality program committees.
\end{IEEEbiography}


\begin{IEEEbiography}
[{\includegraphics[width=1in,height=1.25in,clip,keepaspectratio]{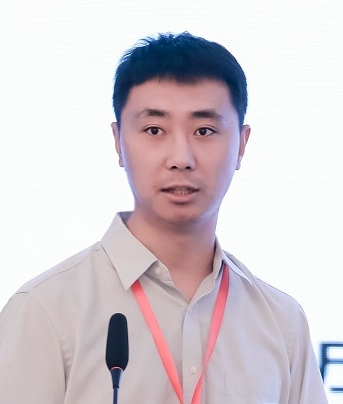}}]{Jiaolong Yang}
is currently a senior researcher at Microsoft Research Asia, Beijing, China. He received the dual Ph.D. degrees in Computer Science and Engineering from the Australian National University and Beijing Institute of Tech nology in 2016. His research interests include 3D vision for human face and body. He serve as the program committee member/reviewer for major computer vision conferences and journals including CVPR/ICCV/ECCV/TPAMI/IJCV, the Area Chair for CVPR/ICCV/ECCV/WACV/MM, and the Associate Editor for the International Journal on Computer Vision (IJCV).
\end{IEEEbiography}


\begin{IEEEbiography}
[{\includegraphics[width=1in,height=1.25in,clip,keepaspectratio]{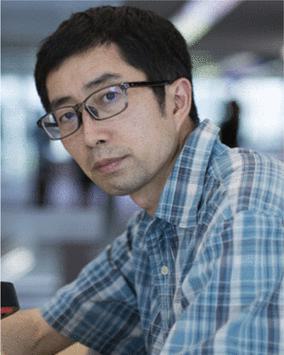}}]{Xin Tong}
received the BS and master's degrees in computer science from Zhejiang University in 1993 and 1996, respectively, and the PhD degree in computer graphics from Tsinghua University in 1999. He is currently a principal researcher with Internet Graphics Group, Microsoft Research Asia. His PhD thesis is about hardware assisted volume rendering. His research interests include appearance modeling and rendering, texture synthesis, and image based modeling and rendering.
\end{IEEEbiography}

\vfill

\end{document}